\documentclass[letterpaper]{article} 
\usepackage{aaai2027}  
\usepackage[hyphens]{url}  
\usepackage{graphicx} 
\usepackage{amsmath} 
\usepackage{amssymb} 
\usepackage{natbib}  
\usepackage{caption} 
\usepackage{algorithm}
\usepackage{algorithmic}
\usepackage{amsmath}
\usepackage{multirow}
\usepackage[table,dvipsnames]{xcolor}
\usepackage{pifont}
\usepackage{newfloat}
\usepackage{listings}
\DeclareCaptionStyle{ruled}{labelfont=normalfont,labelsep=colon,strut=off} 
\floatstyle{ruled}
\newfloat{listing}{tb}{lst}{}
\floatname{listing}{Listing}

\usepackage{booktabs}

\title{MEGA: Object-Level Mesh Extraction from 3D Gaussian Splatting \\ via Spatial Visual Distillation}
\author{
    Liwei Liao\textsuperscript{\rm 1,2},
    Yingkui Zhang\textsuperscript{\rm 2,3},
    Qianqian Tong\textsuperscript{\rm 2},
    Ronggang Wang\textsuperscript{\rm 1,2}\corresponding
}
\affiliations{
    \textsuperscript{\rm 1}Peking University, \textsuperscript{\rm 2}Pengcheng Laboratory, \textsuperscript{\rm 3}Beihang University\\

    levio@pku.edu.cn, rgwang@pkusz.edu.cn
}

\begin{document}
\maketitle

\begin{abstract}
Mesh extraction from 3D Gaussian Splatting (3DGS) aims to endow 3D Gaussians with accurate geometric structures, enabling explicit and precise 3D occupancy. However, existing methods primarily focus on scene-level mesh extraction, making them unable to represent object-level occupancy and often resulting in non-watertight surfaces. To overcome these limitations, we propose \textbf{MEGA} (\underline{M}esh \underline{E}xtraction from \underline{GA}ussians), a ``segment-then-mesh'' framework for extracting object-level, watertight meshes from complex 3DGS scenes.
At the core of MEGA are \textbf{Spatial Visual Distillation (SVD)} and a mask-guided neural surface reconstruction module. SVD treats the 3DGS model as a teacher, sampling diverse camera poses and rendering the corresponding views of each segmented object. These observations are then used to train a mesh reconstruction model through photometric supervision. Extensive experiments on several widely used benchmarks demonstrate that MEGA achieves state-of-the-art performance in recovering accurate object-level 3D occupancy. Moreover, MEGA enables complex physical interactions by combining high-quality object-level meshes for geometric occupancy with 3DGS representations for photorealistic rendering.

\end{abstract}
\section{1. Introduction}

The goal of this work is to extract object-level meshes from complex 3D Gaussian Splatting (3DGS)~\cite{3dgs} scenes, thereby providing explicit and accurate object-level 3D occupancy. As one of the most prominent 3D representations in recent years, 3DGS faces a fundamental limitation: \textit{\textbf{it can be seen, but not touched}}. Although 3DGS explicitly represents a scene using Gaussian primitives, it does not encode coherent surface geometry. Moreover, the anisotropic nature of these primitives means that they do not necessarily lie on the true surfaces of objects. Consequently, 3DGS lacks the geometric structure required for reliable physical interaction, posing a major obstacle to its application in physical-world simulation.
Numerous methods~\cite{guedon2024sugar,huang20242d,chen2024pgsr,guedon2025milo,chang2026meshsplat,held2025meshsplatting,held2026triangle} have attempted to reconstruct meshes from 3DGS, with the goal of providing explicit 3D occupancy and making 3DGS representations ``touchable.'' However, these methods typically suffer from two major limitations. First, they primarily extract scene-level meshes, which are poorly suited to object-level interaction and manipulation. Second, the resulting meshes are often non-watertight and therefore fail to provide complete and reliable 3D occupancy.

\begin{figure*}[t]
    \centering
    \includegraphics[width=\textwidth]{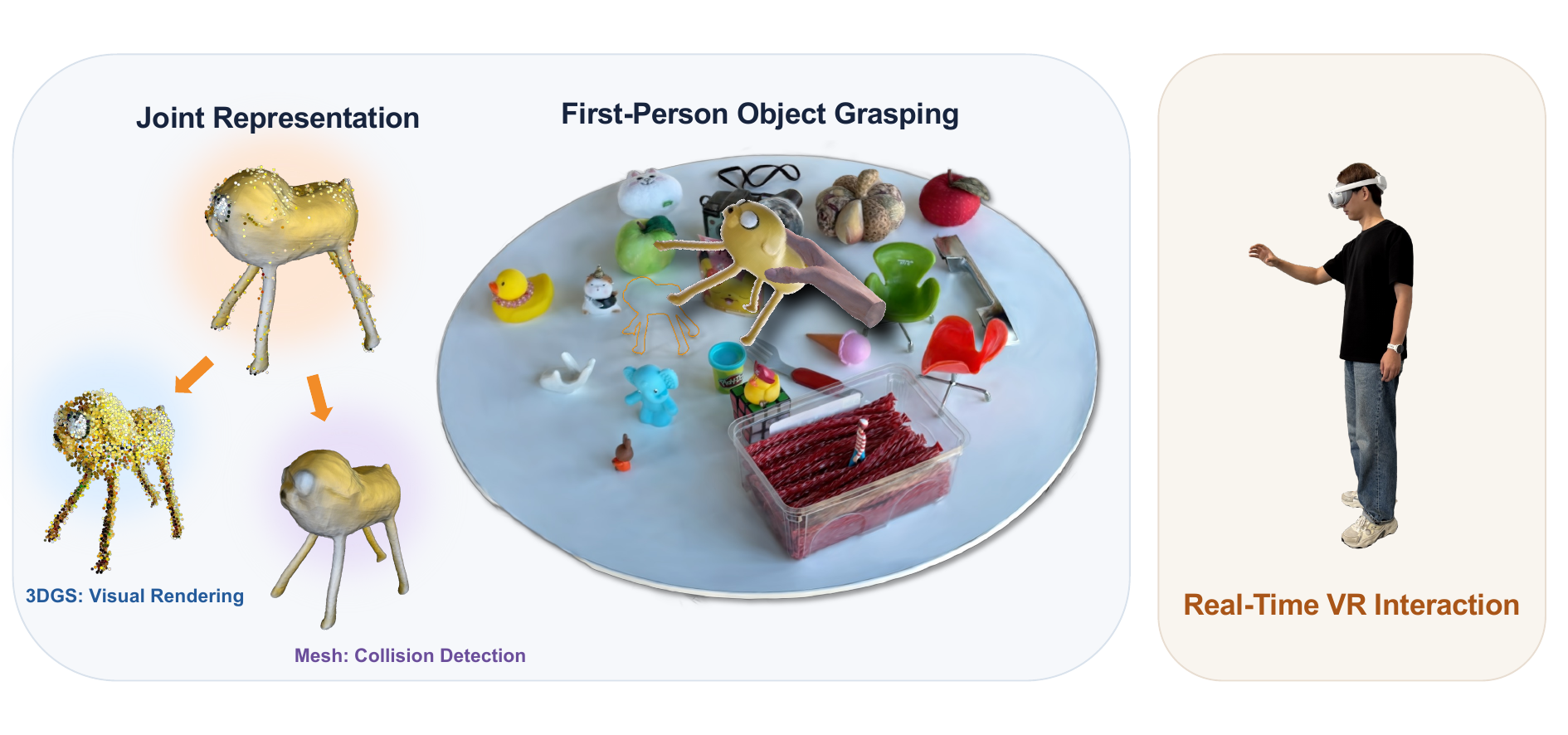}
    \caption{
    MEGA provides object-level 3D occupancy for complex 3DGS scenes by extracting object-level meshes. Each object is jointly represented by an explicit mesh for physical interaction and 3D Gaussians for visual rendering, enabling touch and grasp interactions while preserving high visual quality.
    }
    \label{fig:teaser}
\end{figure*}

\textit{Why, then, is object-level occupancy essential for 3DGS?}

The answer lies in the need to faithfully model the physical world. Although 3DGS excels at representing visual appearance, it lacks the explicit object-level geometry required for physical interaction. Scene-level geometry alone is insufficient: practical applications require individual objects to be localized, manipulated, grasped, and simulated independently. Object-level meshes therefore provide the missing geometric interface between photorealistic 3DGS representations and physically interactive environments.

To bridge this gap, we propose \textbf{MEGA}, a framework for extracting object-level meshes from 3DGS scenes. MEGA not only extracts watertight meshes for individual objects but also maintains accurate geometric and coordinate alignment between the extracted meshes and the underlying 3D Gaussians. It thus provides effective object-level 3D occupancy while supporting richer interactions and manipulations, as illustrated in Fig.~\ref{fig:teaser}.
At the core of MEGA is \textbf{Spatial Visual Distillation (SVD)}. Given a target object, SVD constructs a carefully designed array of virtual cameras and renders the segmented object from diverse viewpoints. These renderings serve as teacher signals for training a neural mesh reconstruction module, ultimately producing an object-level mesh aligned with the coordinate system of the original 3DGS scene. 
In summary, our main contributions are as follows:
\begin{itemize}
    \item We propose \textbf{MEGA}, an object-level mesh extraction framework for 3DGS that provides accurate object-level 3D occupancy in complex scenes and supports rich object-level interactions and manipulations.

    \item We introduce \textbf{Spatial Visual Distillation}, which distills multi-view visual knowledge from 3DGS into a neural mesh reconstruction model. 

    \item We present an early systematic exploration of object-level mesh extraction from 3DGS and establish a strong baseline for future research on physically interactive 3DGS representations.
\end{itemize}


\section{2. RELATED WORKS}

\subsection{2.1. 3DGS Segmentation}

3DGS segmentation partitions the Gaussian primitives in a scene into distinct subsets, each corresponding to an individual object. By assigning semantic identities to otherwise appearance-oriented Gaussian primitives, segmentation provides essential spatial priors for subsequent object-level mesh extraction. Existing methods can be broadly divided into two categories: \textbf{offline methods}, which require per-scene optimization, and \textbf{online methods}, which operate in an inference-only manner.
Among offline methods, SAGA~\cite{cen2025segment} distills scale-gated affinity features from multi-view 2D masks and assigns semantic attributes to individual Gaussians. Click-Gaussian~\cite{choi2024click} aggregates global feature candidates to achieve consistent feature learning across views. COB-GS~\cite{zhang2025cob} further introduces boundary-adaptive Gaussian splitting to resolve ambiguities near object boundaries. Despite their effectiveness, these methods require additional optimization for each scene, which typically takes tens of minutes.

In contrast, online methods perform segmentation without per-scene training. GaussianCut~\cite{jain2024gaussiancut} constructs a graph over Gaussian primitives and formulates segmentation as a graph-partitioning problem. SAGD~\cite{SAGD} introduces Gaussian decomposition to alleviate jagged boundary artifacts. More recently, SAGO~\cite{liao2026online} introduces virtual drones for interactive 3D segmentation, achieving sub-second latency without per-scene optimization. In this work, we adopt SAGO to obtain the object-level segmentation priors required by our framework.

\subsection{2.2. Mesh Extraction from 3DGS}

Numerous methods have explored mesh extraction from 3D Gaussian representations. SuGaR~\cite{guedon2024sugar} introduces a regularization term that encourages Gaussian primitives to align with scene surfaces, followed by mesh extraction using Poisson surface reconstruction~\cite{kazhdan2006poisson}. SAGD~\cite{SAGD} applies the Quickhull algorithm~\cite{barber1996quickhull} to construct collision meshes for segmented objects. However, the resulting meshes are limited to coarse convex polyhedra and cannot accurately capture complex object geometry. 2DGS~\cite{huang20242d} replaces 3D Gaussian primitives with 2D Gaussian disks to improve surface alignment and subsequently extracts meshes through truncated signed distance field (TSDF) fusion. MeshSplatting~\cite{held2025meshsplatting} instead adopts opaque triangles as primitives, enabling the end-to-end reconstruction of explicit meshes.

Despite these advances, most existing methods, with the exception of SAGD, focus primarily on scene-level mesh extraction. Obtaining individual object meshes therefore requires additional manual segmentation or post-processing. Moreover, the extracted meshes are frequently non-watertight, making them unsuitable for representing complete and reliable object-level 3D occupancy. Although SAGD directly produces object-level collision meshes, its convex-hull representation is too coarse to accurately describe object geometry.

These limitations motivate our work. In contrast to existing approaches, MEGA extracts object-level, watertight, and coordinate-aligned meshes from complex 3DGS scenes, providing explicit and accurate object-level 3D occupancy.

\begin{figure*}[t]
  \centering
    \setlength{\abovecaptionskip}{0.3cm}
    \setlength{\belowcaptionskip}{-0.2cm}
  \includegraphics[width=1.0\linewidth]{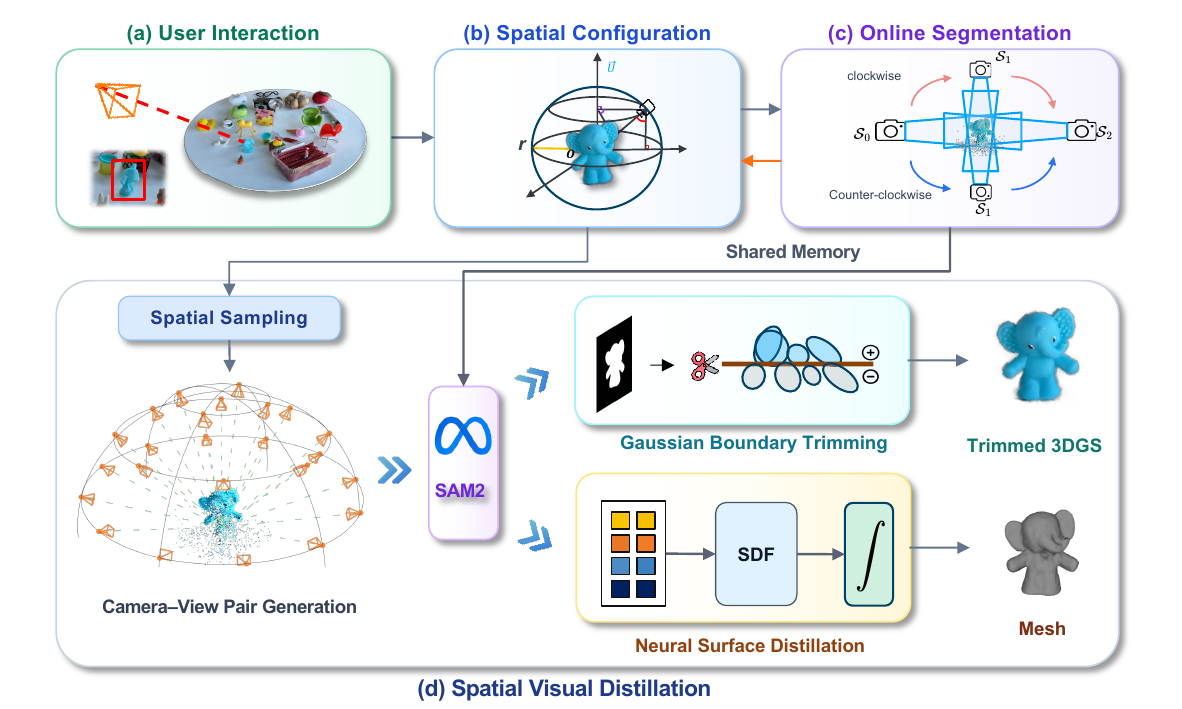}
  \caption{\textbf{Overview of the MEGA framework.}
Given a pretrained 3DGS scene, MEGA proceeds as follows:
(a) the user interactively selects a target object;
(b) the spatial configuration is initialized from the user input;
(c) the online segmentation module identifies the target object and estimates its position, 6-DoF pose, and appropriate observation distance; and
(d) spatial viewpoints are sampled around the object to generate sufficient camera--view pairs, which can optionally be refined using SAM2.
These camera--view pairs subsequently serve as teacher signals for either distilling an object-level mesh or refining boundary artifacts in the object-level 3DGS representation.
  }
  \label{fig:framework}
\end{figure*}

\section{3. Method}

\subsection{3.1. Overview}

We propose \textbf{MEGA}, a framework for extracting object-level meshes from pretrained 3D Gaussian Splatting (3DGS) scenes. Given a pretrained 3DGS representation $\mathbf{\Theta}$ and a user prompt $p$, MEGA identifies the target object and produces both a refined object-level Gaussian representation $\hat{\mathbf{\Theta}}_{\mathrm{obj}}$ and its corresponding mesh $\mathcal{M}_{\mathrm{obj}}$. The complete framework can be expressed as
\begin{equation}
\bigl(
\hat{\mathbf{\Theta}}_{\mathrm{obj}},
\mathcal{M}_{\mathrm{obj}}
\bigr)
=
\operatorname{MEGA}(\mathbf{\Theta};p),
\end{equation}
where $p$ may be a point, a bounding box, or a textual description.

As illustrated in Fig.~\ref{fig:framework}, MEGA follows a ``segment-then-mesh'' paradigm. It first obtains an object-level Gaussian subset using interactive 3DGS segmentation. It then constructs a spatial configuration around the target object and performs \textbf{Spatial Visual Distillation (SVD)} to transfer multi-view visual information from 3DGS to a neural surface representation. Finally, MEGA extracts a watertight mesh from the learned surface and optionally refines boundary artifacts in the object-level Gaussian representation.

\subsection{3.2. Preliminaries: Interactive 3DGS Segmentation}

Interactive 3DGS segmentation allows a user to select a target object through a 2D prompt and subsequently identifies its corresponding Gaussian primitives in 3D space. Formally, the segmentation process partitions the full Gaussian representation $\mathbf{\Theta}$ into foreground and background subsets:
\begin{equation}
\mathbf{\Theta}_{\mathrm{obj}}
=
\operatorname{Segment}(\mathbf{\Theta};p),
\end{equation}
where $p$ denotes the user prompt and $\mathbf{\Theta}_{\mathrm{obj}}$ is the segmented object-level Gaussian subset.

As a segment-then-mesh framework, MEGA requires an efficient 3DGS segmentation method to provide an initial object-level prior. We adopt SAGO~\cite{liao2026online}, an online segmentation framework that introduces virtual drones equipped with SAM2~\cite{ravi2024sam2} trackers and Mask-shaped Frustum Filtering (MFF). SAGO formulates next-best-view (NBV) planning as
\begin{equation}
\mathbf{v}^{*}
=
\arg\max_{\mathbf{v}_{(\Delta\phi,\theta)}}
\left|
\Delta\mathcal{A}'(\mathbf{v})
\right|,
\label{eq:nbv_planning}
\end{equation}
where
\begin{equation}
\Delta\mathcal{A}'(\mathbf{v})
=
\operatorname{MFF}
\left(
\operatorname{SAM2}
\left(
\mathcal{I}(\mathbf{v},\mathcal{A}^{(t)}),
M_t
\right),
\mathcal{A}^{(t)}
\right).
\label{eq:action_of_segmentation}
\end{equation}
Here, $\mathbf{v}_{(\Delta\phi,\theta)}$ denotes a candidate next-best view, $\mathcal{A}^{(t)}$ is the foreground subset at iteration $t$, $M_t$ represents the tracker memory, and $\Delta\mathcal{A}'(\mathbf{v})$ is the predicted update to the foreground subset from viewpoint $\mathbf{v}$. The objective is to select the viewpoint that maximizes the expected foreground update while preserving foreground consistency.

In addition to producing a coarse object segmentation with sub-second latency, SAGO provides useful spatial configuration and tracking-memory priors. We therefore adopt it as the segmentation backbone of MEGA.

\paragraph{Spatial Configuration.}
The spatial configuration $\mathcal{Y}$ defines a spherical coordinate system centered on the target object:
\begin{equation}
\mathcal{Y}
=
\left\{
\vec{\mathbf{U}},
\mathbf{x},
r
\right\},
\end{equation}
where $\vec{\mathbf{U}}$ denotes the ground-plane normal, $\mathbf{x}$ is the centroid of the target object, and $r$ is the preferred observation distance, corresponding to the radius of the sampling sphere. This configuration determines the virtual viewpoints used to generate the multi-view observations for subsequent mesh reconstruction. Additional details about spatial configuration estimation are provided in the appendix.

\subsection{3.3. Spatial Visual Distillation}

We introduce \textbf{Spatial Visual Distillation (SVD)}, a cross-representation distillation paradigm designed for 3D reconstruction. SVD transfers high-quality multi-view observations from a teacher 3DGS representation to a student neural surface representation. This formulation combines the rendering quality of 3DGS with the explicit geometric modeling capability of neural surfaces, ultimately enabling the extraction of an accurate and watertight object-level mesh.

\paragraph{Spatial Sampling.}
Given the segmented Gaussian subset $\mathbf{\Theta}_{\mathrm{obj}}$ and spatial configuration
$\mathcal{Y}=\{\vec{\mathbf{U}},\mathbf{x},r\}$, we construct a set of virtual cameras
\begin{equation}
\mathcal{C}_{\mathrm{cam}}
=
\left\{
c_k
\right\}_{k=1}^{N}
\end{equation}
on the upper hemisphere centered at the object centroid $\mathbf{x}$. Each camera is parameterized in spherical coordinates as
\begin{equation}
c_k=(\phi_k,\theta_k,r),
\end{equation}
where $\phi_k\in[0,2\pi)$ and $\theta_k\in[0,\pi/2]$ denote the yaw and pitch angles, respectively.

The camera orientation is chosen such that its optical axis points toward the object centroid. Specifically, for a camera with center $\mathbf{t}_k$ and rotation matrix $\mathbf{R}_k$, we require
\begin{equation}
\mathbf{R}_k\mathbf{e}_z
=
\frac{
\mathbf{x}-\mathbf{t}_k
}{
\left\|
\mathbf{x}-\mathbf{t}_k
\right\|_2
}.
\end{equation}
The camera intrinsics $\mathbf{K}$ are fixed and shared with the original 3DGS training cameras.

We uniformly sample the yaw and pitch angles according to
\begin{equation}
\phi_i
=
\frac{2\pi i}{N_\phi},
\quad
i=0,\ldots,N_\phi-1,
\end{equation}
and
\begin{equation}
\theta_j
=
\frac{\pi j}{2(N_\theta-1)},
\quad
j=0,\ldots,N_\theta-1.
\end{equation}
This yields a total of
\begin{equation}
N=N_\phi N_\theta
\end{equation}
virtual viewpoints. The resulting hemispherical sampling provides diverse observations of the target object's appearance and geometry, which are essential for high-quality surface reconstruction.

For each virtual camera $c_k$ with extrinsics $(\mathbf{R}_k,\mathbf{t}_k)$ and intrinsics $\mathbf{K}$, we render an RGB image from the segmented object-level Gaussians and obtain its corresponding alpha mask:
\begin{align}
I_k
&=
\mathcal{R}_{\mathrm{RGB}}
\left(
\mathbf{\Theta}_{\mathrm{obj}};c_k
\right),\\
A_k
&=
\mathcal{S}_{\mathrm{SAM2}}
\left(
I_k;\mathbf{M}_{\mathrm{obj}}
\right),
\end{align}
where $\mathcal{R}_{\mathrm{RGB}}$ denotes the differentiable Gaussian rasterizer and $\mathcal{S}_{\mathrm{SAM2}}$ denotes SAM2-based mask propagation using the shared object memory $\mathbf{M}_{\mathrm{obj}}$. The resulting collection
\begin{equation}
\mathcal{D}_{\mathrm{SVD}}
=
\left\{
(I_k,A_k,c_k)
\right\}_{k=1}^{N}
\end{equation}
constitutes the teacher observations used for neural surface distillation.

\paragraph{Neural Surface Distillation.}
We adopt a neural surface reconstruction model based on multi-resolution hash encoding~\cite{muller2022instant}. The target surface is represented by a signed distance function
\begin{equation}
f_{\boldsymbol{\theta}}:
\mathbb{R}^{3}
\rightarrow
\mathbb{R},
\end{equation}
parameterized by $\boldsymbol{\theta}$.

For each sampled camera $c_k$, volume rendering of the neural SDF produces an RGB image $\hat{I}_k$ and an opacity map $\hat{A}_k$:
\begin{equation}
\left(
\hat{I}_k,
\hat{A}_k
\right)
=
\mathcal{R}_{\mathrm{vol}}
\left(
f_{\boldsymbol{\theta}};c_k
\right),
\end{equation}
where $\mathcal{R}_{\mathrm{vol}}$ denotes the volume-rendering operator.

We optimize the neural surface by matching its rendered observations to the teacher views generated from 3DGS:
\begin{equation}
\mathcal{L}_{\mathrm{photo}}
=
\sum_{k=1}^{N}
\left(
\left\|
\hat{I}_k-I_k
\right\|_1
+
\lambda_{\mathrm{mask}}
\left\|
\hat{A}_k-A_k
\right\|_1
\right),
\end{equation}
where $\lambda_{\mathrm{mask}}$ controls the contribution of mask supervision.

To encourage $f_{\boldsymbol{\theta}}$ to represent a valid signed distance field, we additionally impose Eikonal regularization:
\begin{equation}
\mathcal{L}_{\mathrm{eik}}
=
\mathbb{E}_{\mathbf{q}}
\left[
\left(
\left\|
\nabla f_{\boldsymbol{\theta}}(\mathbf{q})
\right\|_2
-1
\right)^2
\right],
\end{equation}
where $\mathbf{q}$ denotes a sampled 3D point. The complete SVD training objective is
\begin{equation}
\mathcal{L}_{\mathrm{SVD}}
=
\mathcal{L}_{\mathrm{photo}}
+
\lambda_{\mathrm{eik}}
\mathcal{L}_{\mathrm{eik}}.
\end{equation}

\paragraph{Mesh Extraction.}
After optimization converges, we evaluate the learned SDF on a dense volumetric grid and apply the marching cubes algorithm to extract an explicit triangular mesh:
\begin{equation}
\mathcal{M}_{\mathrm{obj}}
=
\left(
\mathcal{V}_{\mathrm{obj}},
\mathcal{F}_{\mathrm{obj}}
\right),
\end{equation}
where $\mathcal{V}_{\mathrm{obj}}$ and $\mathcal{F}_{\mathrm{obj}}$ denote the mesh vertices and faces, respectively. Because the neural surface is optimized directly in the coordinate system of the original 3DGS scene, the extracted mesh is spatially aligned with $\mathbf{\Theta}_{\mathrm{obj}}$. The resulting watertight mesh therefore provides explicit and accurate object-level 3D occupancy.

\newcommand{\cmark}{\ensuremath{\checkmark}}
\newcommand{\xmark}{\ensuremath{\times}}

\begin{table*}[h]
\centering
\renewcommand{\arraystretch}{1.5}
\setlength{\tabcolsep}{1.5mm}
\footnotesize

\begin{tabular}{l l c c c c c}
\noalign{\hrule height 1.2pt}

\multirow{2}{*}{Category}
& \multirow{2}{*}{Method}
& \multicolumn{1}{c}{Mesh}
& \multicolumn{2}{c}{SPIn-NeRF}
& \multicolumn{2}{c}{NVOS} \\

\cline{3-3}\cline{4-5}\cline{6-7}

&
& Watertight
& mIoU ($\uparrow$)
& mAcc ($\uparrow$)
& mIoU ($\uparrow$)
& mAcc ($\uparrow$) \\

\hline

\multirow{5}{*}{Gaussians Only}
& NVOS~\cite{ren2022NOVS}
& -- & -- & -- & 70.1 & 92.0 \\

& SAGA~\cite{cen2025segment}
& -- & 93.4 & 99.2 & 92.6 & 98.6 \\

& GaussianCut~\cite{jain2024gaussiancut}
& -- & 92.9 & 99.2 & 92.5 & 98.4 \\

& FlashSplat~\cite{shen2024flashsplat}
& -- & -- & -- & 91.8 & 98.6 \\

& SAGO~\cite{liao2026online}
& -- & 92.5 & 99.3 & 92.7 & 98.7 \\

\hline

\multirow{8}{*}{Meshable}
& SAM3D$^\dagger$~\cite{chen2025sam}
& \ding{51} & 43.2 & 47.1 & 59.2 & 68.2 \\

& SuGaR$^\ddagger$~\cite{guedon2024sugar}
& \ding{55}
& 81.4 & 88.7
& \cellcolor{yellow!25}90.5 & 92.0 \\

& MeshSplatting$^\ddagger$~\cite{held2025meshsplatting}
& \ding{55}
& 76.2 & 78.9 & 81.5 & 89.1 \\

& 2DGS$^\ddagger$~\cite{huang20242d}
& \ding{55}
& 86.8 & 88.9 & 90.1 & 94.2 \\

& PGSR$^\ddagger$~\cite{chen2024pgsr}
& \ding{55}
& 85.6 & 90.9
& \cellcolor{yellow!25}90.5 & 93.7 \\

& \textbf{Our SVD} + 2DGS~\cite{huang20242d}
& \ding{55}
& \cellcolor{orange!25}91.5
& \cellcolor{yellow!25}95.9
& \cellcolor{red!25}91.7
& \cellcolor{yellow!25}96.2 \\

& \textbf{Our SVD} + PGSR~\cite{chen2024pgsr}
& \ding{55}
& \cellcolor{yellow!25}91.1
& \cellcolor{orange!25}96.1
& \cellcolor{yellow!25}90.5
& \cellcolor{orange!25}96.5 \\

& \textbf{Ours (MEGA)}
& \ding{51}
& \cellcolor{red!25}91.8
& \cellcolor{red!25}97.1
& \cellcolor{orange!25}91.4
& \cellcolor{red!25}98.5 \\

\noalign{\hrule height 1.2pt}
\end{tabular}


\caption{
Quantitative comparison with related methods on
SPIn-NeRF and NVOS datasets.
\colorbox{red!25}{1st}, \
\colorbox{orange!25}{2nd}, and
\colorbox{yellow!25}{3rd} best results are highlighted among meshable methods.
The symbol $\dagger$ indicates manual alignment, and
$\ddagger$ indicates manual segmentation.
}
\label{tab:quant}
\end{table*}

\section{4. EXPERIMENTS}
\label{sec:experiments}

\subsection{4.1. Experimental Setting}
In our experiments, we design a Spatial Sampling array consisting of 3 rows, each containing 72 virtual viewpoints, totaling 216 virtual viewpoints for all quantitative evaluations. We employ two small MLPs for Neural Surface Distillation. For detailed information regarding the structure and specifics of the MLPs, please refer to Appendix 1.2. All experiments are conducted on a single NVIDIA RTX 4090 GPU. 

\begin{figure}[t]
  \centering
  \setlength{\abovecaptionskip}{0.3cm}
  \includegraphics[width=1\linewidth]{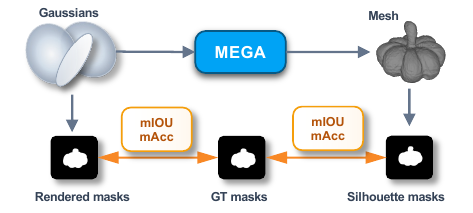}
  \caption{\textbf{Silhouette matching.} The performance of both 3D Gaussian segmentation and object-level mesh extraction can be evaluated within a unified benchmark via silhouettes.
  }
\label{fig:silhouette}
\end{figure}

\paragraph{Evaluation on silhouette Matching.} As shown in Fig.~\ref{fig:silhouette}, to assess the object-level mesh extraction capability, we project the extracted meshes onto the test views and compute mIoU and mAcc against the ground-truth masks. We conduct evaluations on multiple 3D segmentation datasets including LERF-mask~\cite{kerr2023lerf}, NVOS~\cite{ren2022NOVS} and SPIn-NeRF~\cite{mirzaei2023spin}. 

\paragraph{Baselines.} We compare the performance of existing 3D segmentation methods and other mesh extraction methods on this task to validate  MEGA's effectiveness. We compare our method with several state-of-the-art mesh extraction methods, including SAM3D~\cite{chen2025sam}, PGSR~\cite{chen2024pgsr}, MeshSplatting~\cite{held2025meshsplatting}, and 2DGS~\cite{huang20242d}. These methods represent the current best practices in mesh reconstruction and serve as benchmarks for evaluating the performance of our approach.

\begin{figure*}[t]
\centering
\setlength{\abovecaptionskip}{0.3cm}
    \setlength{\belowcaptionskip}{-0.2cm}
\includegraphics[width=0.9\textwidth]{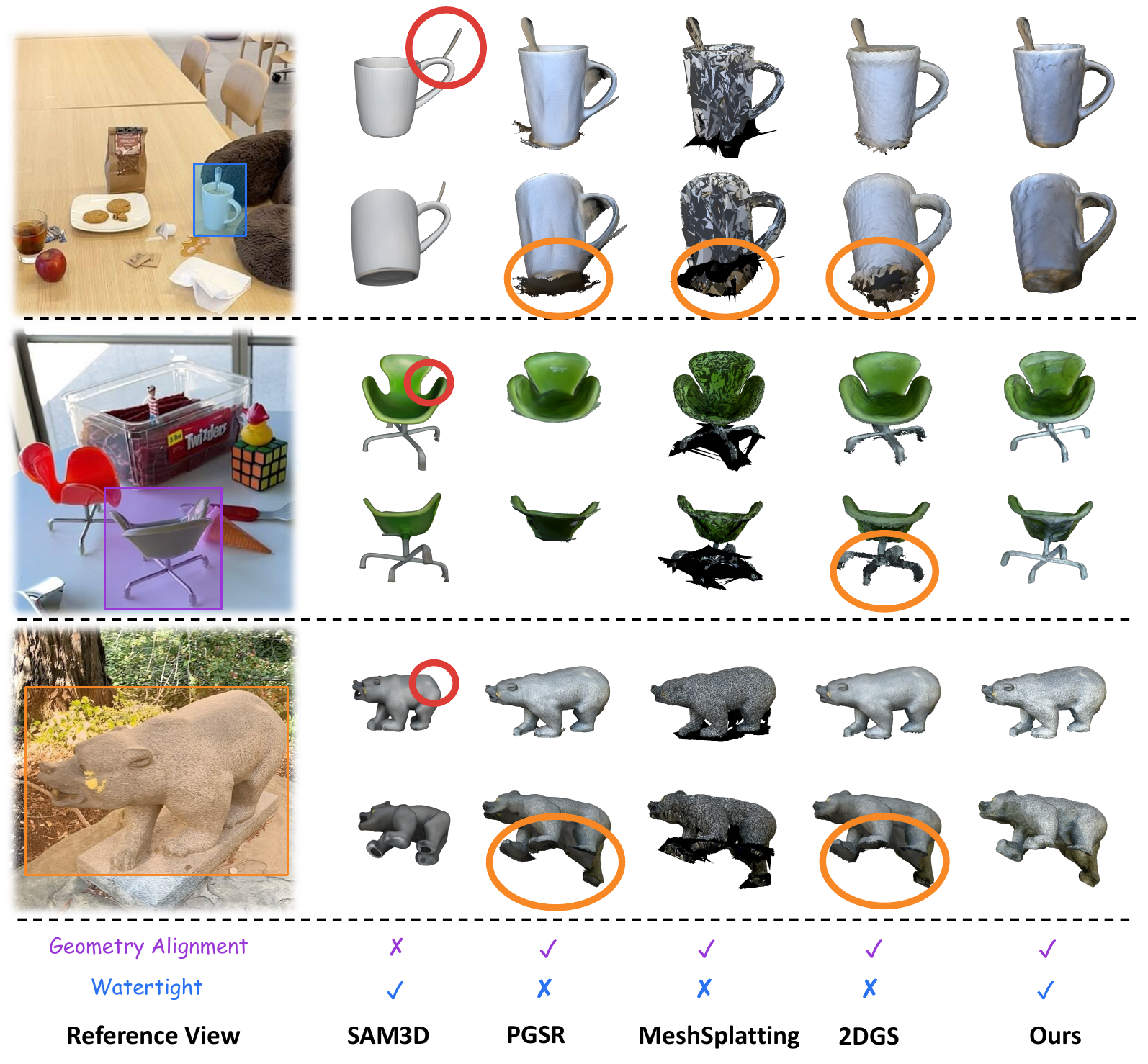}
\caption{The comparison of object-level mesh extraction methods. Red circles indicate the bad alignment while orange circles indicate the holes in the meshes.
}
\label{fig:occ_contrast}
\end{figure*}


\subsection{4.2. Quantitative Evaluation}
As shown in Tab.~\ref{tab:quant}, MEGA achieves state-of-the-art performance in all meshable methods and SVD can work as a plug-and-play module to improve the performance of existing mesh extraction methods. In addition, we also evaluate the performance of MEGA on LERF-Mask dataset~\cite{kerr2023lerf} and the results are shown in Tab.~\ref{tab:lerf_mask}. MEGA achieves competitive performance compared with SAGO, which serves as MEGA's teacher model.

\begin{table}[t]
    \centering
    \setlength{\tabcolsep}{4pt}
    \begin{tabular}{lcccc}
        \noalign{\hrule height 1.2pt}
        Method & Figurines & Ramen & Teatime & Average \\
        \midrule
        LERF             & 33.5          & 28.3          & 49.7          & 37.1          \\
        Gau-Group      & 69.7          & 77.0          & 71.7          & 66.1          \\
        OmniSeg3D     & 87.1          & 77.3          & 73.8          & 79.4          \\
        Click-Gaussian  & 93.2    & \textbf{90.9} & 83.2          & 89.1          \\
        SAGO$^{\S}$         & 93.9 & 90.6          & \textbf{88.6} & \textbf{91.0} \\
        \midrule
        Ours (MEGA)         & \textbf{94.1} & 90.1    & 86.2 & 90.1 \\
        \noalign{\hrule height 1.2pt}
    \end{tabular}
    \caption{Quantitative comparison on LERF-Mask (mIoU $\uparrow$). SAGO$^{\S}$ serves as MEGA's teacher model. Note that SVD can potentially make MEGA exceed its teacher.
    }
    \label{tab:lerf_mask}
\end{table}

\begin{figure*}[t]
\centering
\setlength{\abovecaptionskip}{0.3cm}
    \setlength{\belowcaptionskip}{-0.2cm}
\includegraphics[width=1\textwidth]{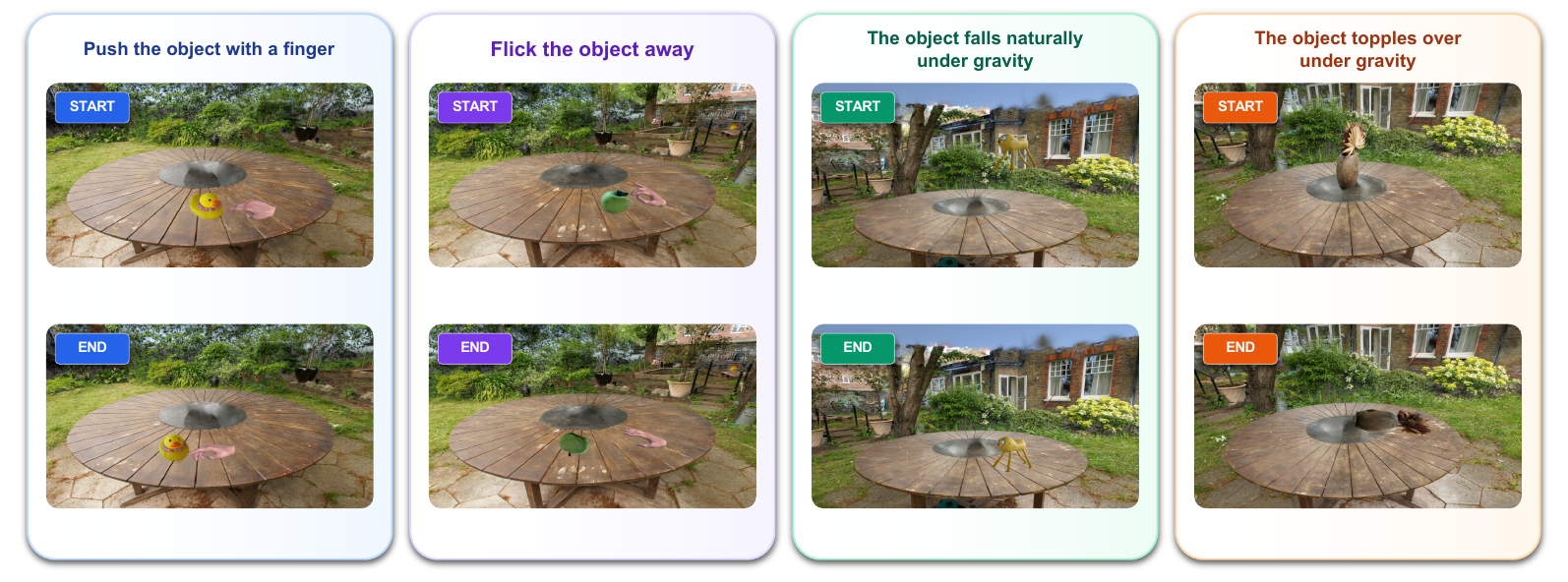}
\caption{We implement object-level Gaussian-Mesh unified representation with MEGA, enabling rich human-object interactions and physics simulations. More examples can be found in the appendix and demo video.
}
\label{fig:physical}
\end{figure*}

\subsection{4.3. Qualitative Comparison}
\begin{figure}[t]
  \centering
  \setlength{\abovecaptionskip}{0.3cm}
  \includegraphics[width=1\linewidth]{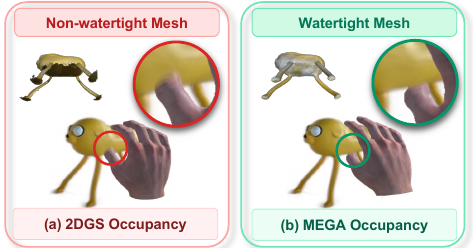}
  \caption{\textbf{Penetration test.} The non-watertight meshes may lead to severe penetration in physical simulations. 
  }
\label{fig:penetration}
\end{figure}

As shown in Fig. ~\ref{fig:occ_contrast}, compared with existing mainstream mesh extraction methods, our MEGA produces more accurate watertight meshes with well-aligned geometric structures. In contrast, although SAM3D can generate a watertight mesh from a single view, its reconstructed geometry deviates substantially from that of the real object. Other splatting-based methods, PGSR, MeshSplatting and 2DGS can all produce geometrically aligned meshes, but they exhibit large holes on the bottom surface, preventing the formation of complete 3D occupancy. The non-watertight meshes may lead to severe penetration in physical simulations, as shown in Fig.~\ref{fig:penetration}. 

\begin{figure}[t]
  \centering
  \setlength{\abovecaptionskip}{0.3cm}
  \includegraphics[width=1\linewidth]{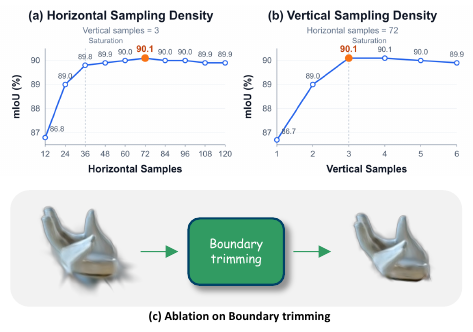}
  \caption{\textbf{Ablation Results} of sampling densification ((a), (b)) and boundary trimming (c).
  }
\label{fig:ablation}
\end{figure}

\subsection{4.4. Applications}
MEGA jointly represents an object using 3DGS and a mesh, allowing physical interaction engines that previously supported only meshes to operate on 3DGS objects. Our preliminary experiments demonstrate at least two potential applications of MEGA: (1) \textbf{Human–Object Interaction} and (2) \textbf{Physics Simulation}. 
We present four examples in Fig.~\ref{fig:physical}, with the first two demonstrating applications in human–object interaction and the latter two showcasing physics simulation. Additional examples can be found in the appendix and demo video. We hope that MEGA will provide more possibilities for the application of 3DGS in various domains.

\subsection{4.5. Ablation Study}

\paragraph{Ablation on Sampling Densification.} We conduct an ablation study on SVD sampling density. Starting from the default setting of 3 rows and 72 columns, we vary the vertical and horizontal densities separately, and then evaluate MEGA’s mIoU on the LERF-Mask dataset. Fig.~\ref{fig:ablation} (a) shows the results of changing the horizontal density while fixing the vertical density, whereas Fig.~\ref{fig:ablation} (b) shows the results of changing the vertical density while fixing the horizontal density. The experiments indicate that once the sampling density reaches a certain level (e.g., 3×72), further increasing the density does not necessarily yield significant gains.

\paragraph{Ablation on Neural Surface Distillation.} We replace the neural surface distillation with existing splatting-based methods by directly training 2DGS and PGSR using the camera-view pairs generated by SVD. As shown in the bottom three rows of Table 1, although the SVD-augmented 2DGS and PGSR underperform the full MEGA in terms of mIoU and mAcc, both still outperform their counterparts trained independently. This demonstrates that SVD can serve as a plug-and-play module to enhance the performance of existing mesh extraction methods.

\paragraph{Boundary Trimming.} Boundary trimming does not affect the mesh extraction results; however, the rough surfaces of 3D Gaussian objects can degrade the visual quality of the Gaussian-Mesh joint representation in MEGA. Therefore, following SAGD~\cite{SAGD} and GaussianTrimmer~\cite{liao2026gaussiantrimmer}, we trim the 3DGS objects to obtain better visual quality, as shown in Fig. 7(c).

\section{5. CONCLUSION}

In this paper, we present MEGA, an object-level mesh extraction framework for 3DGS. By introducing Spatial Visual Distillation (SVD), MEGA uses the segmented 3DGS as a teacher model to generate multi-view camera pairs, which are then used to supervise a neural surface reconstruction network, enabling the extraction of watertight, geometry-aligned object-level meshes for effective 3D occupancy.

\paragraph{Limitations.} Although MEGA effectively extracts meshes for 3DGS objects, the Gaussian fitting may not perfectly align with the surface, leading to slight deviations. As a result, minor mesh penetration may occur during interactions. Future work could further optimize 3DGS using the better-aligned mesh representations to address these issues.

\bibliography{aaai2027}


\end{document}